\PassOptionsToPackage{unicode}{hyperref}
\PassOptionsToPackage{hyphens}{url}
\documentclass[
  american,
]{article}
\usepackage{xcolor}
\usepackage{amsmath,amssymb}
\usepackage{iftex}
\ifPDFTeX
  \usepackage[T1]{fontenc}
  \usepackage[utf8]{inputenc}
  \usepackage{textcomp} 
\else 
  \usepackage{unicode-math} 
  \defaultfontfeatures{Scale=MatchLowercase}
  \defaultfontfeatures[\rmfamily]{Ligatures=TeX,Scale=1}
\fi
\usepackage{lmodern}
\ifPDFTeX\else
\fi
\IfFileExists{upquote.sty}{\usepackage{upquote}}{}
\IfFileExists{microtype.sty}{
  \usepackage[]{microtype}
  \UseMicrotypeSet[protrusion]{basicmath} 
}{}
\makeatletter
\@ifundefined{KOMAClassName}{
  \IfFileExists{parskip.sty}{%
    \usepackage{parskip}
  }{
    \setlength{\parindent}{0pt}
    \setlength{\parskip}{6pt plus 2pt minus 1pt}}
}{
  \KOMAoptions{parskip=half}}
\makeatother
\usepackage{graphicx}
\makeatletter
\newsavebox\pandoc@box
\newcommand*\pandocbounded[1]{
  \sbox\pandoc@box{#1}%
  \Gscale@div\@tempa{\textheight}{\dimexpr\ht\pandoc@box+\dp\pandoc@box\relax}%
  \Gscale@div\@tempb{\linewidth}{\wd\pandoc@box}%
  \ifdim\@tempb\p@<\@tempa\p@\let\@tempa\@tempb\fi
  \ifdim\@tempa\p@<\p@\scalebox{\@tempa}{\usebox\pandoc@box}%
  \else\usebox{\pandoc@box}%
  \fi%
}
\def\fps@figure{htbp}
\makeatother
\NewDocumentCommand\citeproctext{}{}

\makeatletter
 \let\@cite@ofmt\@firstofone
 \def\@biblabel#1{}
 \def\@cite#1#2{{#1\if@tempswa , #2\fi}}
\makeatother
\newlength{\cslhangindent}
\newlength{\csllabelwidth}
\newenvironment{CSLReferences}[2] 
 {\begin{list}{}{%
  \setlength{\itemindent}{0pt}
  \setlength{\leftmargin}{0pt}
  \setlength{\parsep}{0pt}
  \ifodd #1
   \setlength{\leftmargin}{\cslhangindent}
   \setlength{\itemindent}{-1\cslhangindent}
  \fi
  \setlength{\itemsep}{#2\baselineskip}}}
 {\end{list}}
\usepackage{calc}

\ifLuaTeX
\usepackage[bidi=basic,shorthands=off,]{babel}
\else
\usepackage[bidi=default,shorthands=off,]{babel}
\fi
\ifLuaTeX
  \usepackage{selnolig} 
\fi
\usepackage{float}
\usepackage{bookmark}
\IfFileExists{xurl.sty}{\usepackage{xurl}}{} 
\hypersetup{
  pdftitle={Tablassert: Declarative Extraction of Knowledge Graphs from Biomedical Tables},
  pdfauthor={Skye L. Goetz; Jared C. Roach; Gwênlyn Glusman},
  pdflang={en-US},
  hidelinks,
  pdfcreator={LaTeX via pandoc}}

\title{Tablassert: Declarative Extraction of
Knowledge Graphs from Biomedical Tables}
\usepackage{etoolbox}
\makeatletter
\providecommand{\subtitle}[1]{
  \apptocmd{\@title}{\par {\large #1 \par}}{}{}
}
\makeatother
\subtitle{Institute for Systems Biology}
\author{Skye L. Goetz \and Jared C. Roach \and Gwênlyn Glusman}
\date{20 October 2026}

\begin{document}
\maketitle
\begin{abstract}
Tables published in biomedical research papers, particularly
supplemental tables, capture knowledge assertions that may not be in the
main text of the manuscript for several reasons, including low
statistical significance, space limitations, and narrative scope. These
tables include data (e.g., association statistics, assay readouts,
curated gene lists, and omics) encoding relationships between biological
entities. Tablassert converts such tabular data (Excel, CSV, TSV,
Parquet) into knowledge graphs ready for the NCATS Biomedical Data
Translator (Fecho et al. 2025). A user declares in YAML how each table's
data map to subject-predicate-object-qualifier statements; Tablassert
resolves free text (e.g., ``lung cancer'') to standard identifiers,
attaches provenance and statistics, and emits graph files conforming to
Translator standards (D. Unni, Bruskiewich, and Moxon 2026; D. R. Unni
et al. 2022).
\end{abstract}

\section{Statement of Need}\label{statement-of-need}

Biomedical knowledge graphs classically rely on bespoke ingest scripts
for each source. When data sources lack standardization, such as the
millions of tables published in biomedical literature, bespoke scripts
impose three costs that limit scalability. First, mapping each table's
rows, columns, and cells to biological assertions is idiosyncratic,
making them hard to inspect, review, or reuse. Second, entity resolution
may depend on hosted network services, making graph builds
unreproducible. Third, and most importantly, because each source
requires a new pipeline, most published tables are never ingested and
remain neither interoperable nor reusable.

Tablassert supports knowledge-graph engineers and bioinformaticians
working with tabular biomedical sources who need standards-conformant
output. Declarative YAML recipes make data-to-statement mapping
scalable, inspectable, and reusable. Batched resolution against an
embedded identifier database removes runtime service dependencies. These
recipes extend beyond journal supplemental tables to any relational
source (trial registries, drug approval records, or a researcher's own
tables). Each build can emit a standardized resource descriptor, like
the Resource Ingest Guide metadata required for integration with
Translator.

\section{State of the Field}\label{state-of-the-field}

Several tools address aspects of biomedical knowledge graph compilation.
KGX (D. Unni, Bruskiewich, and Moxon 2026) exchanges and validates
graphs, including schema-driven Biolink conformance checks, but provides
no column-to-statement mapping, study-level structural invariants,
offline Translator infores checks, or in-place pruning. LinkML (Moxon et
al. 2025) models and validates schemas but does not perform entity
resolution. Koza (Schaper et al. 2026), the Monarch Initiative's (Putman
et al. 2024) ingest engine, pairs YAML configuration with
semi-declarative Python transforms, so complex sources often require
repetitive procedural code and external mappings for resolution.
BioCypher (Lobentanzer et al. 2023) standardizes biomedical graph
construction, and OntoWeaver (Dreo et al. 2024) adds declarative tabular
mappings. However, Translator Biolink qualifier semantics and Resource
Ingest Guide metadata fall outside their scope. SPIRES (Caufield et al.
2024) extracts assertions from free text using zero-shot prompting
rather than semi-structured tables. HarmoMed (Villanueva-Miranda et al.
2026) uses agentic AI to create unified cross-dataset views instead of
reasoning-ready knowledge assertions. Finally, the hosted SRI
NameResolution service (The Biomedical Data Translator Consortium 2026a)
and Node Normalization service (The Biomedical Data Translator
Consortium 2026b) provide identifier resolution and canonical records
through network APIs. Their need for hosting makes them dependent on
service availability, and makes historical results hard to reproduce.

We built Tablassert rather than contributing to an existing project for
three reasons. First, Tablassert's recipe model is not
backward-compatible with the translator-ingests transform interface
(NCATS Biomedical Data Translator Consortium 2025); migrating would
discard provenance, certain edge types, and several qualifiers
attributed to each edge. Second, Tablassert targets many heterogeneous
tables, making additions easy by editing simple recipes rather than
writing new code. Third, deterministic Tablassert builds run locally on
demand, whereas shared ingests run as periodic batch jobs with
unpinnable dependencies, slowing iteration and harming reproducibility.
Together, these choices define Tablassert as a local, declarative,
reproducible table-to-graph system.

\section{Software Design}\label{software-design}

\begin{figure}
\centering
\pandocbounded{\includegraphics[keepaspectratio,alt={Tablassert Architecture. Graph and table YAMLs encode declarative transformations that are validated and coerced, compiled into operations, lazily optimized, and executed using a Python-Rust runtime, embedded named-entity recognition, and cryptographic caching. Outputs from individual tables are aggregated into one Biolink-conformant knowledge graph, exported as KGX nodes and edges alongside Translator Resource Ingest Guide (RIG) metadata.}]{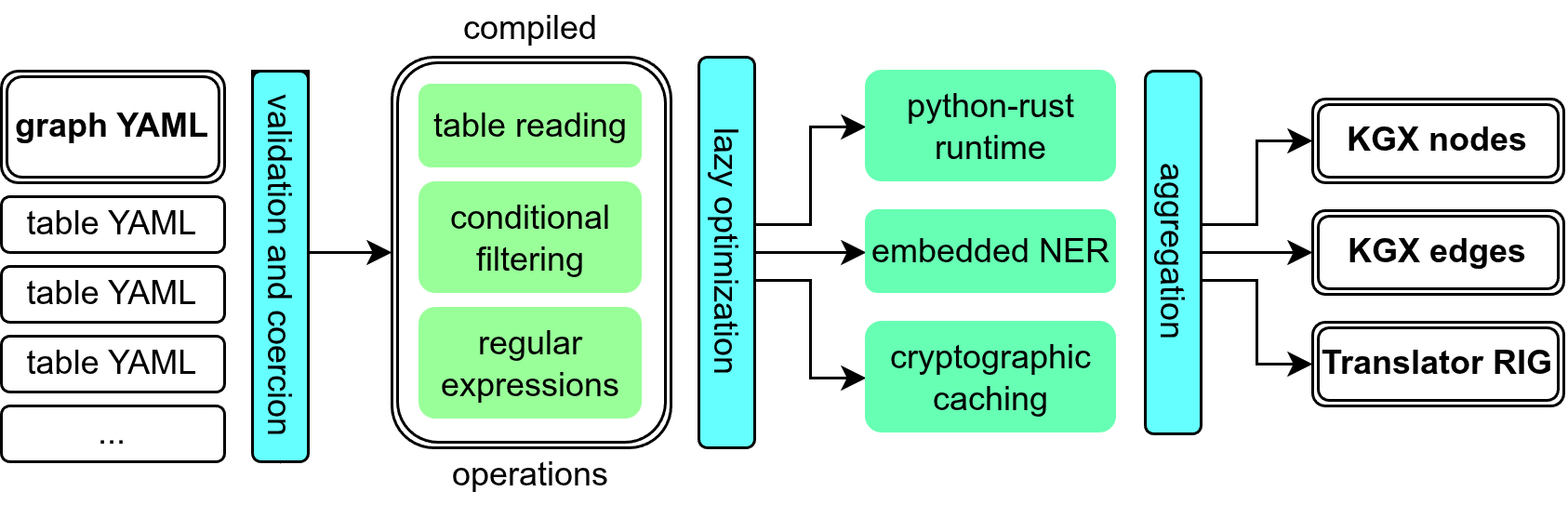}}
\caption{\textbf{Tablassert Architecture.} Graph and table YAMLs encode
declarative transformations that are validated and coerced, compiled
into operations, lazily optimized, and executed using a Python-Rust
runtime, embedded named-entity recognition, and cryptographic caching.
Outputs from individual tables are aggregated into one
Biolink-conformant knowledge graph, exported as KGX nodes and edges
alongside Translator Resource Ingest Guide (RIG)
metadata.\label{tablassert}}
\end{figure}

Tablassert uses declarative configuration to transform tabular data into
one aggregated, Biolink-conformant knowledge graph exported as KGX with
RIG metadata (Figure \ref{tablassert}). It has three main trade-offs: an
embedded NoSQL (fullmap) database enables local batched resolution
without service calls; declarative YAML favors accessibility,
validation, and speed over procedural flexibility; and a hybrid
Python-Rust Maturin build system favors performance over implementation
simplicity.

\subsection{Embedded fullmap database}\label{embedded-fullmap-database}

Built from BABEL (Morris et al. 2026) exports, the embedded fullmap maps
normalized synonyms and cross-vocabulary meta-identifiers to canonical
identifiers with category and taxon metadata. Synonyms and queries share
normalization. Builds rely on Redb (Berner 2026) for parallel,
read-only, memory-bounded point lookups without SQL or runtime service
calls.

Storage is the main cost: the current BABEL 22.2.0 archive is 44.5 GB
compressed, while the 2026-07-22 redb store benchmarked here occupies
92.9 GB. Resolution uses exact matches over normalized keys. To limit
storage and lookup cost, the current build retains the top 100
experimental taxa by Translator knowledge-graph node frequency, all
OrganismTaxon rows, and rows without taxon metadata, potentially
excluding rare species-specific identifiers. Sixteen smaller shards
accelerate lookups and support multithreaded reads. After the initial
build, read-only access needs no writer coordination, and a bounded
in-process cache reuses hits and misses with minimal memory and
synchronization costs.

Node columns declare an NCBI taxon filter that retains matches for the
declared species and taxon-less rows, Biolink category priority and
avoid lists, and CURIE-prefix and regex exclusions. Before lookup, the
cleaning trims raw preferred names and synonyms, strips surrounding
quotes, and lowercases. A first normalization stems ASCII tokens,
removes duplicates, and sorts remaining tokens; matching is thus
unaffected by word repetition and order. A second pass removes non-word
characters. Candidates are ranked in this order: exact preferred-name
agreement, normalized-name agreement, category priority, and
normalization level required for the match, with ties retained in the
best tier. Optional column-category frequencies provide a heuristic
tie-breaker.

Optional quality control applies a four-stage cascade: exact or alias
matching, fuzzy string similarity, Schwartz-Hearst abbreviation
expansion (Schwartz and Hearst 2003), then SapBERT (Liu et al. 2021;
RENCI Named Entity Recognition Working Group 2025) embedding similarity
using a checkpoint fine-tuned on BABEL synonyms. Each stage handles only
what cheaper stages could not clear, and suspect mappings are dropped
and logged with scores. For example, an audit rejected ``factor ii''
mapped to \texttt{HMDB:HMDB0000607} (cyanocobalamin; SapBERT cosine
0.444): BABEL lists ``Factor II'' as its synonym, but the coagulation
drug context calls for prothrombin, not vitamin B12 (The UniProt
Consortium 2026).

\subsection{Declarative YAML
configuration}\label{declarative-yaml-configuration}

YAML recipes declare source tables, transformations, and statement
mappings instead of code. Pydantic models validate these recipes before
execution with field-specific errors. The non-Turing-complete DSL favors
common, fast transformations over arbitrary computation, so complex
cross-table mappings and non-Polars-compatible regular expressions fall
outside its scope. This reduces code changes needed for new sources and
helps nontechnical contributors and AI agents author graphs.

Recipes compile into TCode, an ordered sequence of (callable, arguments)
operations executed through Polars lazy frames. Independent
transformations fuse until batched resolution, quality control, or
output writing requires materialization. Sections are statement mappings
that inherit a template's shared source, provenance, metadata, and
qualifier declarations. For example, a section can select another entity
column and predicate while reusing everything else, avoiding a second
nearly identical recipe. Sections that share the same instructions reuse
snapshots from a cryptographic cache. Lazy optimization decouples recipe
order from execution order. It often improves performance but can be
suboptimal for some workloads and rules out transformations requiring
strict procedural ordering.

Builds emit deduplicated KGX nodes and can emit a Resource Ingest Guide.
Edges carry recipe-declared structured provenance, supporting-study
metadata, and statistical qualifiers, including source URLs and
publications when provided; traceability depends on the source and
configuration.

An optional study stage checks structural invariants;
\texttt{validate-kgx} checks category-specific Biolink records and
Translator infores identity offline. The \texttt{-\/-prune} option
removes failing edges and gates the post-prune exit status. These
validations establish structure, not biological validity.

\subsection{Maturin build system}\label{maturin-build-system}

Tablassert's Python layer handles recipe parsing, validation, lazy
execution planning, and build coordination, while compiled Rust supplies
repeated low-level operations. PyO3, a Rust-Python bridge, exposes those
Rust functions as an ordinary Python extension. For example, Rust
normalizes terms, queries the sharded fullmap, streams database
construction through sorted temporary files, and writes KGX output as
NDJSON, a line-oriented JSON format.

CPython's global interpreter lock (GIL) allows only one thread to
execute Python bytecode at a time. Rust calls release the GIL and use
Rayon to spread work across CPU cores on standard CPython (trading build
complexity for multicore performance). Maturin packages the extension as
Linux and macOS PyPI wheels, so users can install them without a Rust
toolchain. However, Windows requires WSL or a source build.

\section{Research Impact Statement}\label{research-impact-statement}

Tablassert underlies several production knowledge graphs. MicrobiomeKG
(Goetz, Glen, and Glusman 2025) was first built with an early
version and its current deployment and maintenance builds use
Tablassert. Multiomics Knowledge Provider (MOKP) (Goetz and
Glusman 2026d) uses Tablassert. The Drug Approvals Knowledge Provider
(DAKP), originally developed in the Airflow-based glusman-team/dakp
repository (Goetz and Glusman 2026a) and now maintained in
glusman-team/graph-monorepo (Goetz and Glusman 2026c),
publishes validated, versioned drug-approval graphs with per-edge
evidence linkouts (Goetz and Glusman 2026b) with Tablassert. As of release,
Multiomics KP 3.1.3 contains 222,994 nodes and 18,206,965 edges derived
from 2,136 PubMed Central (PMC) publications (Goetz and
Glusman 2026d), and DAKP 1.28.1 contains 24,008 nodes and 164,662 edges
(Goetz and Glusman 2026b). KRAKEN (Glen et al. 2026), Phenome
Health's unified knowledge graph, integrates MicrobiomeKG, MOKP, and
DAKP as domain-specific sources. Translator ingest integration is in
progress, beginning with a Multiomics passthrough (Goetz
2026a, 2026b). Future directions entail expanding MOKP to integrate a
comprehensive subset of rare disease supplemental tables. Tablassert's
generality will also support knowledge extraction in other fields,
including Synthetic Chemistry, Materials Science, and the Geosciences.

\section{AI Usage Disclosure}\label{ai-usage-disclosure}

Generative AI assisted with code generation, test writing, security
analyses, documentation, reviewing, drafting, and revising. Coding-agent
use began in January 2026. Models used include claude-opus-5,
claude-sonnet-5, deepseek-v4, gemini-3.8-flash, glm-5.3, gpt-5.6, gpt-6,
kimi-k2, kimi-k3, muse-spark-1.3-contributor, nemotron-3-ultra,
laguna-s-2.1, qwen3.8, mimo-v2.5, and mimo-v2.6-pro in a custom Pi
coding agent harness (v0.81.1) with prompt engineering from promptfoo
(v0.121.17) and GEPA (v0.1.4). AI evaluation with CodeRabbit
supplemented human code review. Grammarly also revised documentation and
manuscripts. The authors made all architecture decisions, reviewed all
outputs, and retain full responsibility for the software and manuscript.

\section{Author Contributions}\label{author-contributions}

Skye Goetz developed the software, led design, and performed analyses.
Jared Roach helped design the software and analyses, and provided
feedback. Gwênlyn Glusman conceptualized the project, helped design the
software and analyses, and provided feedback. All authors wrote the
manuscript.

\section{Acknowledgements}\label{acknowledgements}

This work was supported by the National Science Foundation (Award
2150265) and by the National Center for Advancing Translational
Sciences, Biomedical Translator Program (Other Transaction Awards
OT2TR003443 and OT2TR005706). Any opinions expressed in this manuscript
are those of the Translator community at large and do not necessarily
reflect the views of NCATS, individual Translator team members, or
affiliated organizations and institutions.

NCATS provided financial support but had no role in software design,
analysis, or manuscript preparation.

\section*{References}\label{references}
\addcontentsline{toc}{section}{References}

\protect\phantomsection\label{refs}

\end{document}